\documentclass{article}

    \PassOptionsToPackage{round}{natbib}

    \usepackage[final]{neurips_2020}

\usepackage[utf8]{inputenc} 
\usepackage[T1]{fontenc}    
\usepackage{hyperref}       
\usepackage{url}            
\usepackage{booktabs}       
\usepackage{amsfonts}       
\usepackage{nicefrac}       
\usepackage{microtype}      

\usepackage{todonotes}
\usepackage{amsmath}
\usepackage{amssymb}
\usepackage{bm}
\usepackage[normalem]{ulem}
\usepackage{caption}
\usepackage{subcaption}
\usepackage{amsthm}
\usepackage{wrapfig}
\usepackage[export]{adjustbox}

\newcommand{\stkout}[1]{\ifmmode\text{\sout{\ensuremath{#1}}}\else\sout{#1}\fi}

\def\eqref#1{equation~\ref{#1}}

\def\1{\bm{1}}

\def\eps{{\epsilon}}

\DeclareMathAlphabet{\mathsfit}{\encodingdefault}{\sfdefault}{m}{sl}
\SetMathAlphabet{\mathsfit}{bold}{\encodingdefault}{\sfdefault}{bx}{n}

\def\gA{{\mathcal{A}}}

\def\gD{{\mathcal{D}}}
\def\gE{{\mathcal{E}}}

\def\gL{{\mathcal{L}}}
\def\gM{{\mathcal{M}}}
\def\gN{{\mathcal{N}}}
\def\gO{{\mathcal{O}}}

\def\gS{{\mathcal{S}}}

\def\sR{{\mathbb{R}}}

\newcommand{\E}{\mathbb{E}}

\newcommand{\KL}{D_{\mathrm{KL}}}

\DeclareMathOperator{\Tr}{Tr}

\theoremstyle{definition}
\newtheorem{remark}{Remark}[section]

\theoremstyle{definition}
\newtheorem{definition}{Definition}[section]

\theoremstyle{plain}
\newtheorem{proposition}{Proposition}[section]

\title{A Geometric Perspective  on\\Self-Supervised Policy Adaptation}

\author{%
  Cristian Bodnar\thanks{Work done as a research intern at Google.} \\
  University of Cambridge \\
  Cambridge, UK \\
  \texttt{cb2015@cam.ac.uk} \\
   \And
   Karol Hausman \\
   Robotics at Google \\
   Mountain View, California \\
   \texttt{karolhausman@google.com} \\
   \And
   Gabriel Dulac-Arnold \\
   Google Research \\
   Paris, France \\
   \texttt{dulacarnold@google.com} \\
   \And
   Rico Jonschkowski \\
   Robotics at Google \\
   Mountain View, California \\
   \texttt{rjon@google.com} \\
}

\begin{document}

\maketitle

\begin{abstract}
One of the most challenging aspects of real-world reinforcement learning (RL) is the multitude of unpredictable and ever-changing distractions that could divert an agent from what was tasked to do in its training environment. While an agent could learn from reward signals to ignore them, the complexity of the real-world can make rewards hard to acquire, or, at best, extremely sparse. A recent class of self-supervised methods have shown promise that reward-free adaptation under challenging distractions is possible. However, previous work focused on a short one-episode adaptation setting. In this paper, we consider a long-term adaptation setup that is more akin to the specifics of the real-world and propose a geometric perspective on self-supervised adaptation. We empirically describe the processes that take place in the embedding space during this adaptation process, reveal some of its undesirable effects on performance and show how they can be eliminated. Moreover, we theoretically study how actor-based and actor-free agents can further generalise to the target environment by manipulating the geometry of the manifolds described by the actor and critic functions. 
\end{abstract}

\section{Introduction}

Real-world environments are characterised by an ever-changing set of distractions such as modifications in lighting conditions, object colour variations or evolving backgrounds that are irrelevant for the tasks RL agents should perform. These distractions are often so complex and diverse that they cannot all be anticipated at training time. While further RL training in the target environment could address this problem, RL is based on a reward signal, which usually requires instrumentation or manual labelling. Another way to address the problem of changing distractions is to have the agent continuously adapt to them -- without requiring reward -- in a self-supervised manner.

\citet{hansen2020self} have made important progress in this direction. They propose an agent that implicitly adjust its state representations by training an inverse dynamics model that predicts actions from pairs of states. This network is pre-trained in the source environment and then fine-tuned in the target environment that includes the distractions, which through the shared state representation improves RL performance is the target domain. While their work opened up this exciting avenue of research, the authors mostly focused on a one-episode adaptation process for a Soft Actor-Critic (SAC) \citep{haarnoja2018soft} agent. 

In this work, we consider a long-term reward-free adaptation scenario both in an actor-critic and actor-free setting and provide a geometric description of the processes that take place in the embedding space during the adaptation phase. Firstly, we demonstrate that while the two environments move towards each other in the embedding space, the original representation of the source environment that the agent was trained on is altered. To address this problem, we propose a parallel training procedure which adjusts the actor and critic weights to compensate for the changes in the state representations. Secondly, we formulate an upper bound on the mismatch between the actions taken between the two environments and show how this can be reduced in practice by manipulating the geometry of the manifold described by actor and critic functions.   

\section{Background}

\paragraph{Problem Statement.} We consider two Partially Observable Markov Decision Processes (POMDPs) \citep{astrom1965optimal, KAELBLING199899} $\gM_1 = (\gO, \gS, \gA, T, R, \Omega_1, \gamma)$ and $\gM_2 = (\gO, \gS, \gA, T, R, \Omega_2, \gamma)$ sharing the same observation space $\gO$, state space $\gS$, action space $\gA$, transition function $T(s' | s, a)$, reward function $R(s, a)$, and discount factor $\gamma$, but with distinct conditional observation densities $\Omega_1(o | s, a)$ and $\Omega_2(o | s, a)$, respectively. $\gM_1$ represents the source environment the agent is trained in and $\gM_2$ represents the target (adaptation) environment the agent is deployed in. Because we are interested in reward-free adaptation in the target environment, we assume we do not have access to the reward function $R$ when interacting with environment $\gM_2$. We define our objective as maximizing the expected total reward $\E\big[\sum_{t=0}^\infty \gamma^t R(s_t, a_t)\big]$ in the target environment $\gM_2$.

\begin{wrapfigure}{r}{.5\textwidth}
\includegraphics[width=0.49\linewidth]{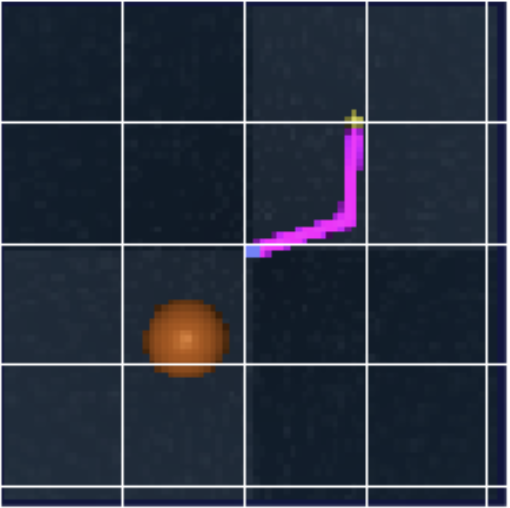}
\includegraphics[width=0.49\linewidth]{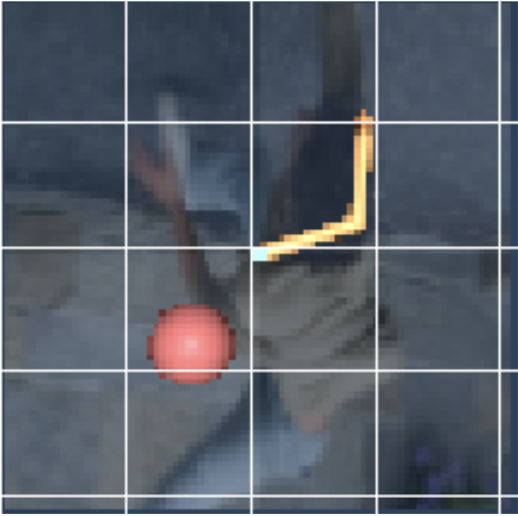}
\caption{Color (left) and background (right) distraction environments for the task reacher-easy.}
\label{fig:envs}
\vskip -0.2in
\end{wrapfigure}

\paragraph{Environments.} For our experiments, we consider the Distracting Control Suite~\citep{distracting_control} based on DM Control~\citep{tassa2018deepmind}. We proceed by training in the distraction-free DM Control environments and consider two different environments for adaptation: video backgrounds and random colour changes. In the video background environment, for each episode, a random frame from a set of 10 videos is used in the background. In the colour distraction environment, the colours of all objects are uniformly sampled from the original colour $\pm0.5$ in each episode. One frame from each of the environments can be seen in Figure \ref{fig:envs}.

\paragraph{Self-Supervised Adaptation} \citet{hansen2020self}\ consider a soft-actor critic (SAC) \citep{haarnoja2018soft} model with an auxiliary inverse dynamics loss. The actor, the critic and the inverse dynamics prediction network share a convolutional encoder. At training time, the whole model is trained in the source domain using the usual SAC loss combined with an auxiliary inverse dynamics prediction loss. At deployment time, the SAC objective is dropped and the agent is adapted to the target domain by minimizing only the inverse dynamics loss for the pairs of consecutive observations it encounters in the target environment. The gradients of this loss are propagated only through the inverse dynamics network and the common encoder. The actor and the critic are left untouched by this adaptation procedure. 

Fine-tuning the encoder for inverse dynamics prediction at testing time in this way was empirically shown to allow the encoder representations to adapt to the distractions present in the target environment. Ultimately, this improves RL performance in that environment without ever having access to the reward signal. In this work, we analyse ways the adaptation process could be improved, while maintaining the same training procedure as in \citet{hansen2020self} in the source environment. Additionally, we extend our analysis to QT-Opt \citep{kalashnikov2018qt}, an actor-free algorithm extensively used in robotic applications.   

\section{Geometry of Self-Supervised Adaptation}

\textbf{Preliminaries.} For our analysis and experiments, we are interested in adapting both in an actor-critic and in an actor-free setting. For the first setting, we use Soft Actor-Critic (SAC) \citep{haarnoja2018soft}, a popular choice in model-free RL. For the latter, we use QT-Opt \citep{kalashnikov2018qt}, a Q-Learning \citep{watkins1992q} based algorithm whose real-world generalization in robotic applications has been well demonstrated \citep{kalashnikov2018qt, bodnar2019quantile}. 

\begin{figure}[ht]
    \centering
    \includegraphics[width=0.7\textwidth]{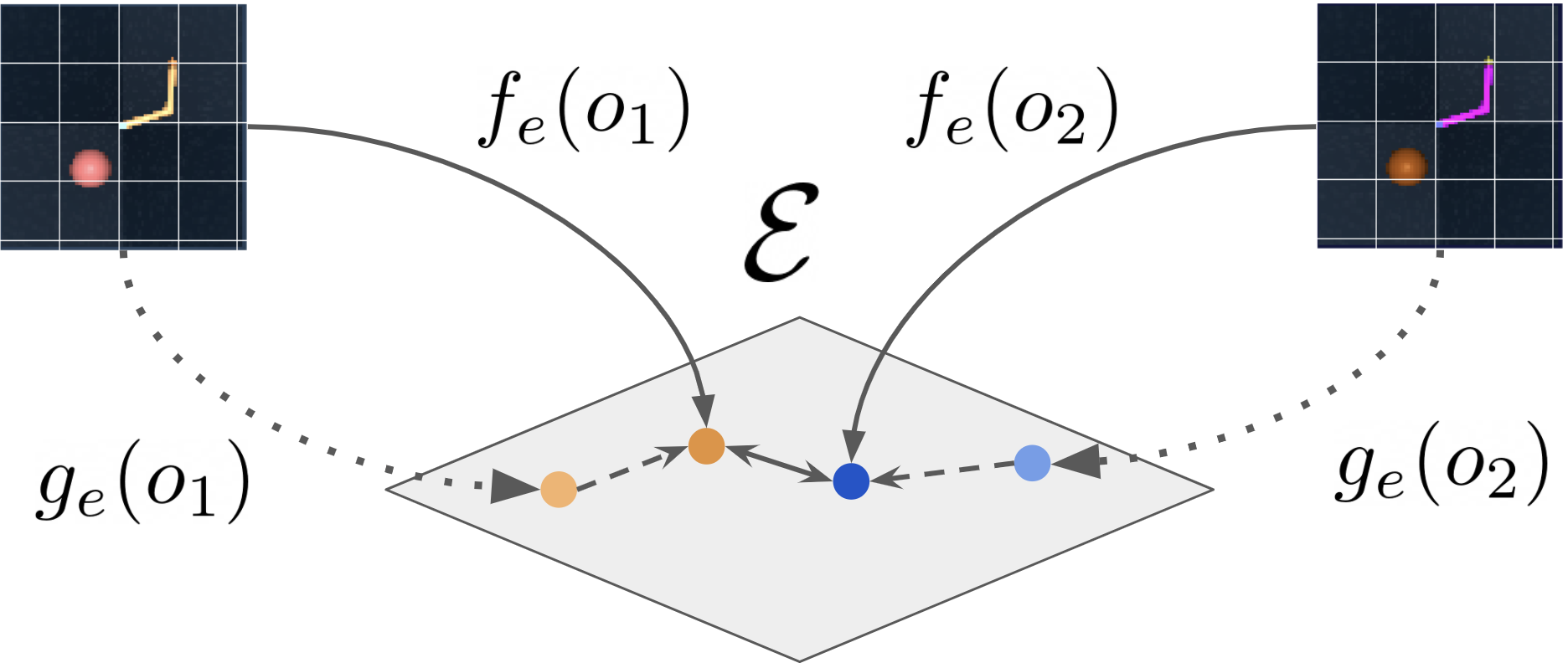}
    \caption{The evolution of the representations of two matching observations during adaptation. The dotted lines indicate the original position of the representations before adaptation ($g_e(o_1)$ and $g_e(o_2)$). The dashed lines indicate how they move in the embedding space and approach each other. }
    \label{fig:diagram}
\end{figure}

In our experiments, we use the same neural network architecture based on \citet{yarats2019improving}: We employ an encoder $f_e: \gO \to \gE$ with eight convolutional layers and ReLU activations that maps from the observation space to the embedding space $\gE$. The embedding space is treated as a proxy for the latent state space of the environment. The encoder, is shared by three similar neural network heads $f_i: \gE \times \gE \to \gA$ (inverse dynamics), $f_c: \gE \times \gA \to \sR$ (critic), $f_a: \gE \to \gA$ (actor -- used only in SAC). Each of the heads is formed of 4 more convolutional layers with the last containing a layer normalised \citep{ba2016layer} and tanh-activated bottleneck of dimension $100$. The bottleneck is followed by two more ReLU activated hidden layers with $1000$ neurons each and an output layer of the corresponding dimension for each function. We will refer to the bottleneck activations of the actor as $b_a: \gE \to \sR^{100}$. Additionally, we use $g$ to refer to state of the networks before adaptation is started (i.e. $g_e$, $g_i$, $g_a$, $g_c$). 

We train our model in the source environment for $250$ thousand steps using two random crop augmentations per state like in DrQ~\citep{kostrikov2020image} and \citet{hansen2020self}. Then, we adapt in the target environment for $50,000$ transitions, with one gradient step per frame. We use a batch size of $512 = 64 \times 8$ containing 64 states with eight random crop augmentations for each. We adapt from a replay buffer with capacity $50,000$ that is filled initially with $3,200$ transitions collected by the trained policy.  

\subsection{The embedding space dynamics}
\label{sec:how_adapt_works}

In this section, we analyse the dynamics in the embedding space for the representations of the two environments in an attempt to elucidate the internal mechanisms of self-supervised adaptation process introduced by \citet{hansen2020self}. Firstly, we hypothesise that the improvements in the total reward brought by the adaptation process are caused by the fact that the representations of the two environments become more similar during adaptation. 

To validate this, we measure the expected distance between the embeddings of observations sharing the same underlying state, formally given by $\E_{s, a}[d(f_e(o_1), f_e(o_2))]$, with $o_1 \sim \Omega_1(s, a)$ and $o_2 \sim \Omega_2(s, a)$. To do so, we collect five matching episodes in the two environments by synchronising the initial state of the two and taking the same actions in both of them. Here, we choose $d$ to be the cosine distance. 

\begin{figure}[ht]
     \centering
     \vspace{5pt}
     \begin{adjustbox}{minipage=\linewidth,scale=0.8}
     \begin{subfigure}[b]{0.48\columnwidth}
         \centering
         \includegraphics[width=\textwidth]{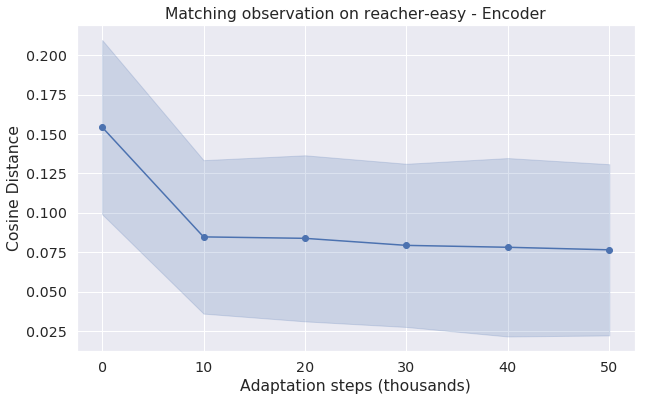}
     \end{subfigure}
     \hfill
     \begin{subfigure}[b]{0.48\columnwidth}
         \centering
         \includegraphics[width=\textwidth]{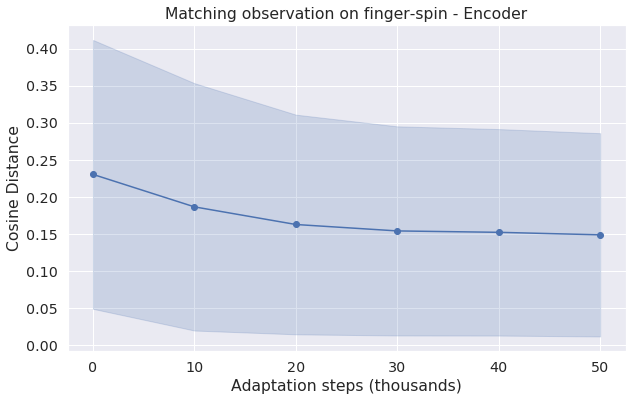}
     \end{subfigure}
     \end{adjustbox}
     \caption{Distance between emeddings of matching states in the reacher-easy and finger-spin environments. The matching states of the two environments move towards each other during adaptation. Lines show means and shaded areas cover two standard deviations of the mean estimate.}
     \label{fig:match_dist_cos}
\end{figure}

In Figure \ref{fig:match_dist_cos} we plot this average distance as a function of the adaptation step for the reacher-easy and finger-spin environments with colour distractions. It shows that the auxiliary loss minimization during the adaptation process implicitly minimizes the distance between matching observations of the two environments. Consequently, this allows the agent trained on the source environment to generalize to the target environment. It remains to be examined as part of future work what types of auxiliary objectives implicitly produce a better alignment of the features and how feature alignment could be perhaps explicitly optimised for.  

However, even though the two environments move closer to each other in the embedding space as we have just shown, we further hypothesise that the original representations of the source environment are progressively forgotten. A large perturbation in the original representations would cause catastrophic forgetting of the actions learned in the source environment, which would likely propagate to the actions taken in the target environment. 

\begin{figure}[h]
     \centering
     \begin{adjustbox}{minipage=\linewidth,scale=1.0}
     \begin{subfigure}[b]{0.32\textwidth}
         \centering
         \includegraphics[width=\textwidth]{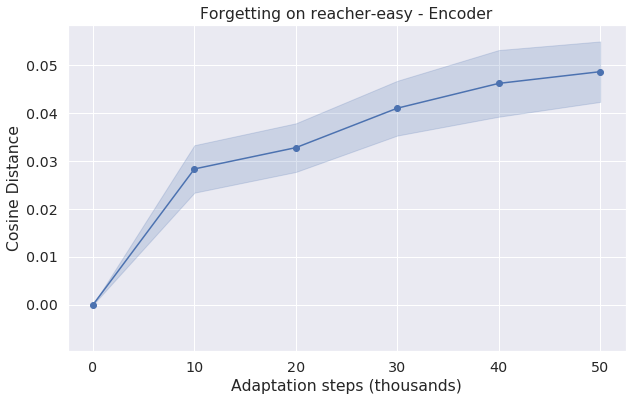}
     \end{subfigure}
     \hfill
     \begin{subfigure}[b]{0.32\textwidth}
         \centering
         \includegraphics[width=\textwidth]{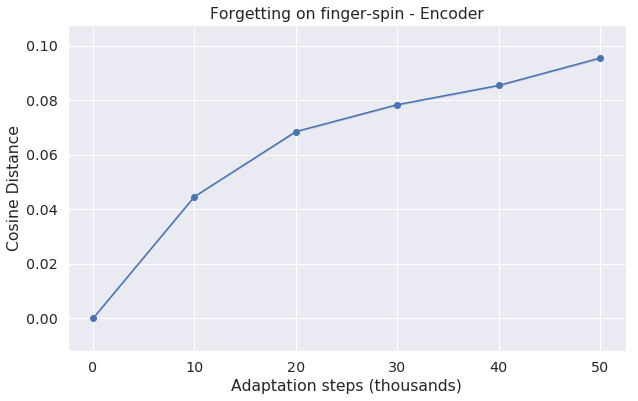}
     \end{subfigure}
     \hfill
     \begin{subfigure}[b]{0.32\textwidth}
         \centering
         \includegraphics[width=\textwidth]{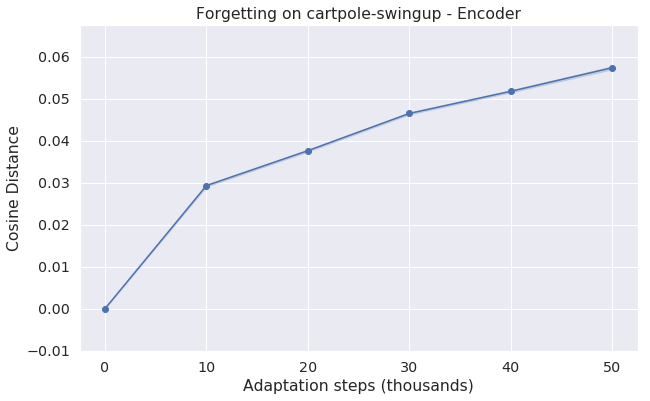}
     \end{subfigure}
     \end{adjustbox}
     \hfill
     \caption{The distance between the embeddings at adaptation time $t$ and the same embeddings at time $t_0$. The original representations are progressively forgotten during adaptation. Note: the last two plots have an extremely small standard deviation across episodes.}
     \label{fig:forgetting}
\end{figure}

To quantify this forgetting, we measure the expected cosine distance between the representations of a set of source observations before adaptation and the representations of the same observations at a later time in the adaptation process. We plot the evolution of these distances during adaptation in Figure \ref{fig:forgetting} for three of the environments. We see that the cosine distance monotonically increases during adaptation, meaning that the original policy is gradually forgotten. In turn, this directly affects the performance of the policy in the source environment during adaptation and the total reward decreases in all source environments as shown by the blue line in Figure \ref{fig:sac_source}. 

\subsection{Bounding the action-mismatch}
\label{sec:theory}

We visually summarize these findings in the diagrammatic illustration in Figure \ref{fig:diagram} for a pair of matching observations of the source and target environments. Starting from this model, in this section, we perform a theoretical analysis of self-supervised adaptation.  

Let $g_e: \gO \to \gE$ be the state of the encoder $f_e$ before adaptation. Then, based on the previous results, we expect the embeddings of two matching observations $o_1$ and $o_2$ to be at some distance $\eps_e = d(g_e(o_1), f_e(o_2))$ away from each other. This distance would depend on how much forgetting has taken place and how close to each other the two environments have become. At the same time, we would expect the action mismatch between the two environments to increase with this distance. In what follows, we formalise these intuitions.    


\begin{definition}[Lipschitz continuous function \citep{o2006metric}]
Given two metric spaces $(X, d_x)$ and $(Y, d_y)$, a function $f: X \to Y$ is K-Lipschitz continuous if there exists a constant $K \geq 0$ such that $d_y(f(x_1), f(x_2)) \leq K d_x(x_1, x_2)$ for all $x_1$ and $x_2$. We refer to the smallest such $K$ as the Lipschitz constant of the function $f$. 
\end{definition}

For this paper, we use the usual Euclidean distance as the metric associated with the domain and co-domain of the functions. The proofs of the results can be found in Appendix \ref{sec:proofs}. 

\begin{proposition}
\label{prop:sac_bound}
Let $f_a(e) = [\mu(e), \sigma^2(e)]$ be the components of $f_a$ that specify the mean and variance of the multivariate (normal) action distribution of the SAC actor. Additionally, let $\mu$ and $\sigma^2$ be $K$-Lipschitz continuous and $\sigma^2_i \geq \sigma^2_{min}$ for all components $i$. Let $e_1$ and $e_2$ be the embeddings of two matching observations with $d(e_1, e_2) = \eps_e$. Then we have that $\KL(\gN(\mu(e_1), \sigma(e_1)) || \gN(\mu(e_2), \sigma(e_2)) \in O((\eps_e K)^2)$. 
\end{proposition}

This proposition formalises the intuition that the closer the two states are and the smoother the actor function is, the more similar the two action distributions are going to be. 

Obtaining a similar bound for QT-Opt is more challenging. Because the actions are selected through a maximisation operation $\arg\max_a Q(s, a)$, any potential bound on the action mismatch would depend on the landscape of $Q(s, \cdot)$. This is stated formally in the following proposition.
\begin{proposition}
\label{prop:qtopt_order}
Let $|f_c(e_1, a_1) - f_c(e_1, a_2)| = \Delta_1$ be the predicted Q-value difference for actions $a_1$ and $a_2$ at observation embedding $e_1$ with $f_c(e_1, a_1) > f_c(e_1, a_2)$. Let $e_2$ be the embedding of another observation. Assume we have a metric over $\gE \times \gA$ with the property that $d([e_1, a_1], [e_2, a_1]) = d([e_1, a_2], [e_2, a_2]) = d(e_1, e_2) = \eps_e$. Additionally, let $f_c$ be $K$-Lipschitz continuous. Then if $\eps_e < \Delta_1/ (2K)$, the order between predicted $Q$ values at $e_2$ is preserved and we have $f_c(e_2, a_1) > f_c(e_2, a_2)$. 
\end{proposition}

This result says the order between any two predicted $Q$ values can be preserved for embeddings in an open ball of radius $\Delta_1/ (2K)$ centred at $e_1$. To increase the size of this ball, we would like to maximise its radius. First thought would be to increase the value of $f_c(e_1, a_1)$ as much as possible and decrease the other $Q$ values in order to increase $\Delta_1$. However, $\Delta_1$ also depends on $K$ and cannot be arbitrarily increased.  

\begin{proposition}
\label{prop:qtopt_unimodal}
Let $d$ be a metric with $d([e_1, a_1], [e_1, a_2]) = d([e_2, a_1], [e_2, a_2]) = d(a_1, a_2) = \eps_a$. Then $\Delta_1 / (2K)$ can be at most $\frac{\eps_a}{2}$. 
\end{proposition}

This shows that the best we could do for a $K$-Lipschitz critic function is to have a unimodal landscape, where the $Q$ values of other actions strictly decrease with the distance from the optimal action. 

From the perspective of Lipschitz continuity, these results describe how one can manipulate the policy's behaviour in the target environment by exploiting the ``stiffness'' of the manifold produced by the actor or critic functions, where the ``stiffness'' is given by the Lipschitz constant. While a low Lipschitz constant gives more power to control the behaviour in the target environment, it can affect the performance in the source environment if the actor and critic functions are not flexible enough. Therefore, these trade-offs must be carefully considered. 

From a distance minimization perspective, it is clear that one should try to reduce $d(g_e(o_1), f_e(o_2))$ as much as possible to reduce the action mismatch between the two embeddings. To that end, we can use the following remark. 
\begin{remark}
From the triangle inequality we have that 
\begin{equation}
d(g_e(o_1), f_e(o_2)) \leq d(g_e(o_1), f_e(o_1)) + d(f_e(o_1), f_e(o_2))
\end{equation}
\end{remark}
This explicitly upper bounds $d(g_e(o_1), f_e(o_2))$ on the amount of forgetting that has taken place (the first term) and how well the auxiliary objective has brought the two environments closer to each other (the second term). As shown in Section \ref{sec:how_adapt_works}, the inverse dynamics objective of \citet{hansen2020self} implicitly minimizes the second term and undesirably increases the first. Therefore, one would like to keep $d(g_e(o_1), f_e(o_1))$ as close to zero as possible. However, this could interfere with the self-supervised objective. In the next section, we propose a better alternative that allows us to consider only $d(f_e(o_1), f_e(o_2))$. 

\subsection{Behaviour Cloning-Based Adaptation}

\begin{figure}
    \centering
    \includegraphics[width=0.8\textwidth]{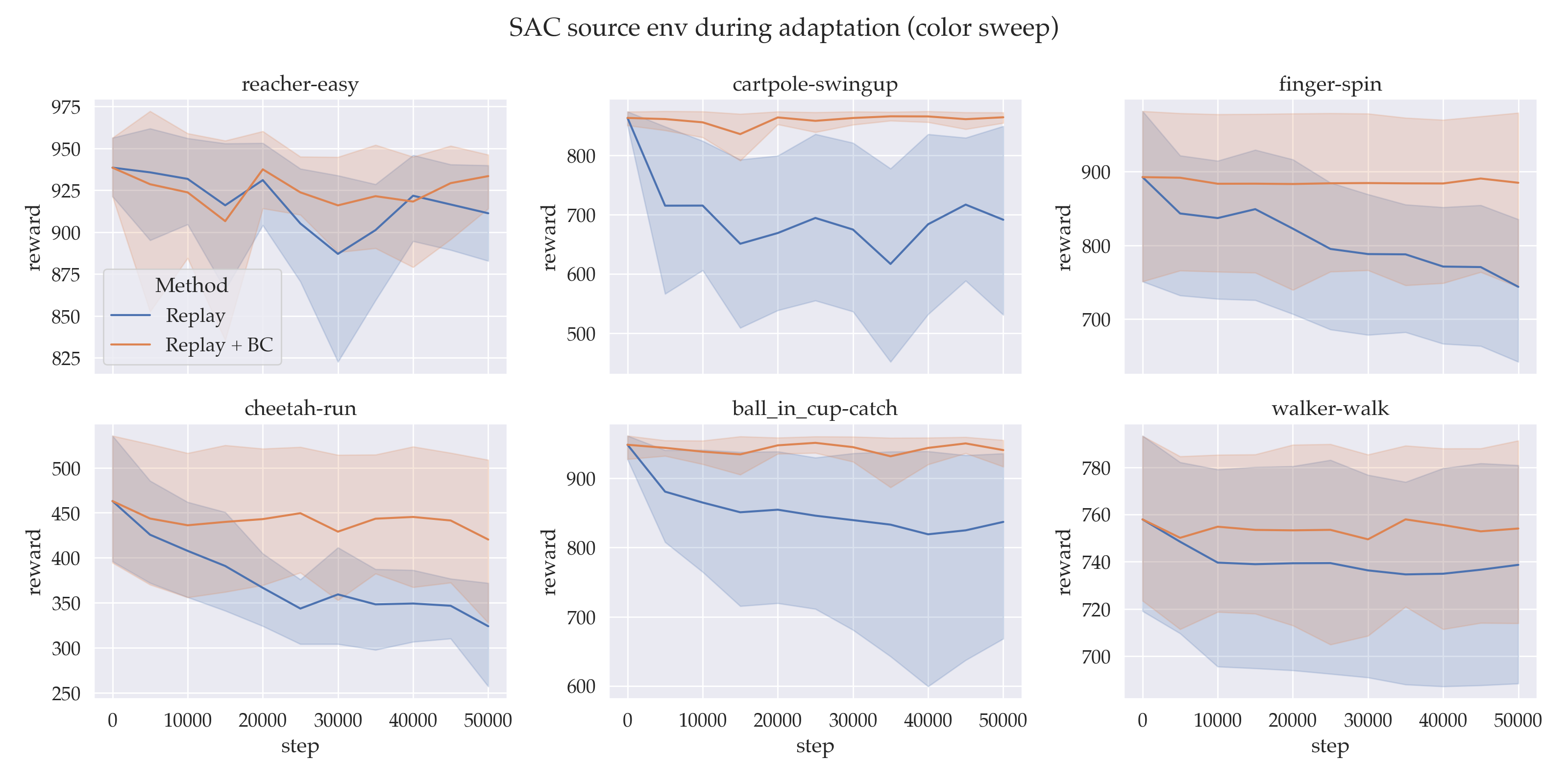}
    \caption{SAC performance in the source environments during adaptation in the color sweep environments. Behaviour cloning stops the catastrophic forgetting of the policy obtained from the training phase and performance stays constant.}
    \label{fig:sac_source}
\end{figure}

\paragraph{Method Overview.} To address the catastrophic forgetting problem, we consider a parallel data collection strategy together with a loss split across the two environments. The loss combines the self-supervised objective in the target environment with a behaviour cloning loss in the source environment, which ensures that $f_a(g_e(o)) \approx f_a(f_e(o))$ for SAC and $f_c(g_e(o)) \approx f_c(f_e(o))$ for QT-Opt, even though $g_e(o) \neq f_e(o)$. Therefore, the action mismatch would depend approximately only on $d(f_e(o_1), f_e(o_2))$. 

\paragraph{SAC.} For SAC, we clone the weights of the encoder and the actor into networks $g_e = f_e$ and $g_a = f_a$ before adaptation. Then, at adaption time, we use $g_a(g_e)$ as a target action to approximate on states coming from the source environment. The gradient of this loss is propagated only through the actor network. Concurrently, we continue minimizing the inverse dynamics loss as before, with gradients propagated through the inverse dynamics network and the encoder. Ultimately, this results in the following loss
\begin{align*}
    \gL = \E_{o \sim \gD_1}\big[\KL[g_a(g_e(o)) || f_a(\bar{f}_e(o) ]\big] + \E_{(o_t, a_t, o_{t+1}) \sim \gD_2}\big[ (f_i(f_e(o_t), f_e(o_{t+1})) - a)^2) \big],
\end{align*}
where $\gD_1$ and $\gD_2$ represent the replay buffer for environments $\gM_1$ and $\gM_2$ and $\bar{f}_e$ denotes the gradients are stopped from propagating through the encoder. The loss makes the actor adjust to the changes in the original representation to preserve its original behaviour. 

\paragraph{QT-Opt.} Similarly, for QT-Opt, we use the target encoder and target critic network from the training stage $f_e^T$, $f_c^T$ and use it as a target for the $Q$ values of the state-action pairs on the source environment. As with SAC, we backpropagate this additional loss only through $f_c$, but not through $f_e$. We obtain a similar loss function:
\begin{align*}
    \gL = \E_{o, a \sim \gD_1}\big[(f_c(\bar{f}_e(o), a) - f_c^T(f_e^T(o), a))^2  \big] 
    + \E_{(o_t, a, o_{t+1}) \sim \gD_2}\big[(f_i(f_e(o_t), f_e(o_{t+1})) - a)^2) \big] 
\end{align*}
This loss makes the critic $f_c$ adjust its weights to compensate for the adjustment in the representations and predict the same Q-values in the source environment.

\section{Results}

We now describe our results for the behaviour cloning-based method proposed in the previous section, as well as for adjusting the Lipschitz constant of the actor-critic functions. 

\subsection{Behaviour Cloning}

We compare the proposed method with an online adaptation process \citep{hansen2020self}, which adapts only on the latest collected transition from the target environment with multiple crop augmentations. Another baseline is the replay buffer-based adaptation previously described in the experimental section. Additionally, we include for reference the original performance of the agent before adaptation and an agent trained normally using rewards in the target environment. 

\begin{figure}[ht]
    \centering
    \vspace{5pt}
    \includegraphics[width=0.8\textwidth]{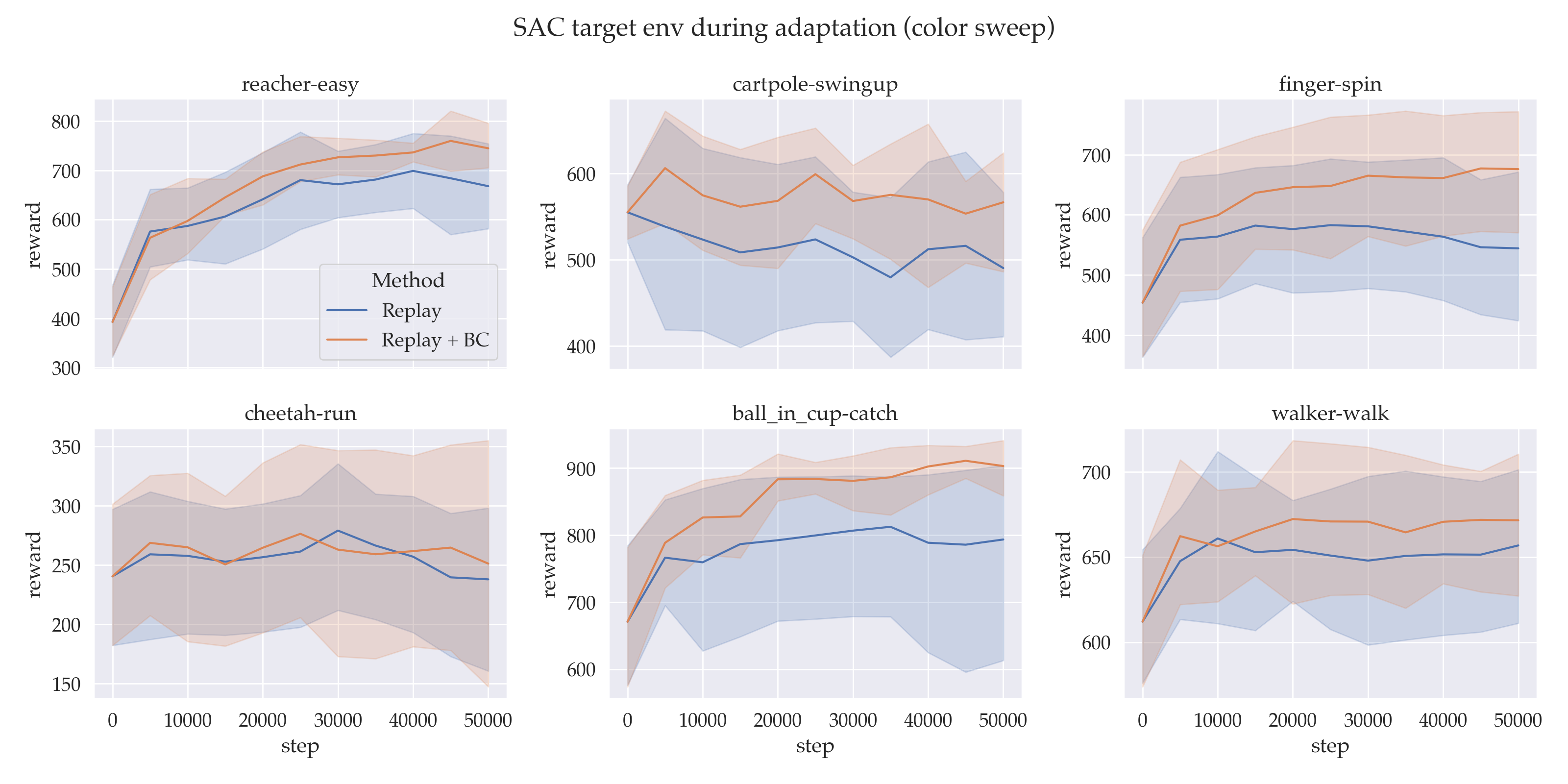}
    \caption{SAC adaptation results for the color sweep environment. Behaviour cloning improves the adaptation performance across all environments.}
    \label{fig:sac_adapt_color}
    \vspace{-10pt}
\end{figure}

As shown by Figure \ref{fig:sac_source}, for SAC, the behaviour cloning loss completely prevents the degradation in performance observed in the vanilla model. In turn, this translates to an improvement in the vast majority of environments, as shown in Tables \ref{tab:sac_adapt_color} and \ref{tab:sac_adapt_bg} (Appendix \ref{sec:extra_adapt}). This is also depicted graphically for the color distraction environments in Figure \ref{fig:sac_adapt_color}. For QT-Opt, the catastrophic forgetting is significantly attenuated in most environments, but not completely reduced because even tiny differences in the $Q$ values can make the maximisation step select another action (Appendix \ref{sec:extra_source}). Again, this translates in improvements in the target environments, as indicated by Tables \ref{tab:qtopt_adapt_color} and \ref{tab:qtopt_adapt_bg} in Appendix \ref{sec:extra_adapt}. More figures describing can be found in Appendix \ref{sec:extra_adapt}. 

At the same time, we notice that the online adaptation intensifies the forgetting process when adapting for multiple episodes and performance consequently degrades. While the replay-buffer based adaptation works better, forgetting still happens. 

\begin{table}[h]
    \centering
    \caption{SAC adaptation in the color distraction environments. Behaviour cloning (BC) either performs similarly or better than the vanilla adaptation.}
    \resizebox{\columnwidth}{!}{
    \begin{tabular}{l | cccccc}
        \toprule
        Method & Reacher-easy & Cartpole-swingup & Finger-spin & Cheetah-run & Ball-in-cup catch & Walker-Walk   \\
        \midrule
        Online & 182.06 $\pm$ 65.56  &  93.52 $\pm$ 18.30 & 136.75 $\pm$ 49.43 & 87.73 $\pm$ 35.87 & 271.63 $\pm$ 150.27 & 331.90 $\pm$ 83.07 \\
        Replay & 667.94 $\pm$ 52.65  &  490.51 $\pm$ 50.30 & 544.24 $\pm$ 73.62 & \textbf{237.92 $\pm$ 40.43} & 793.65 $\pm$ 88.79 & \textbf{656.78 $\pm$ 26.73} \\
        Replay + BC     & \textbf{744.74 $\pm$ 26.50} & \textbf{566.80 $\pm$ 42.09} & \textbf{676.01 $\pm$ 59.60} & \textbf{251.19 $\pm$ 59.48} & \textbf{903.19 $\pm$ 23.40} & \textbf{671.48 $\pm$ 24.47} \\
        \midrule
        RL on Target & 210.04 $\pm$ 27.57  & 605.23 $\pm$ 12.49 & 719.59 $\pm$ 57.72 & 310.79 $\pm$ 40.69 & 777.85 $\pm$ 36.27 & 616.82 $\pm$ 48.30 \\
        Original & 392.64 $\pm$ 42.37 & 555.23 $\pm$ 18.23 & 453.86 $\pm$ 58.35 & 240.46 $\pm$ 33.46 & 670.99 $\pm$ 61.49 & 612.17 $\pm$ 21.96 \\
        \bottomrule
    \end{tabular}
    }
    \label{tab:sac_adapt_color}
\end{table}

\subsection{Distances in the embedding space revisited}

The result from Proposition \ref{prop:sac_bound} relies on a Euclidean norm to find an upper bound on the KL divergence. However, our earlier results from Section \ref{sec:how_adapt_works} have used another metric: the cosine distance. Therefore, motivated by our results, we look again at the distance between the embedding by using a mean squared error (MSE) metric. 

\begin{figure}[h]
     \centering
     \begin{adjustbox}{minipage=\linewidth,scale=0.7}
     \begin{subfigure}[b]{0.48\columnwidth}
         \centering
         \includegraphics[width=\textwidth]{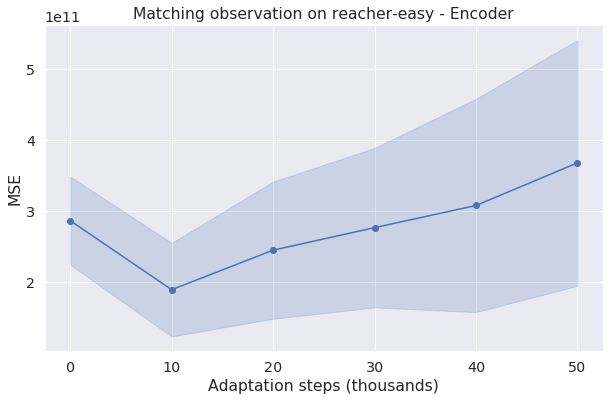}
     \end{subfigure}
     \hfill
     \begin{subfigure}[b]{0.48\columnwidth}
         \centering
         \includegraphics[width=\textwidth]{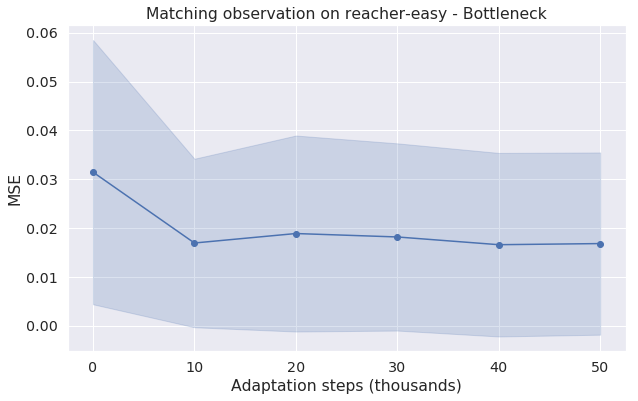}
     \end{subfigure}
     \end{adjustbox}
     \caption{Distance between emeddings of matching states computed at the encoder level (top) and bottleneck level (bottom). Only the bottleneck embeddings display the desired behaviour: a low-magnitude decreasing distance.}
     \label{fig:dist_mse}
\end{figure}

As shown by Figure \ref{fig:dist_mse} (left), when using an MSE metric, perhaps surprisingly, the distance has a huge magnitude of $10^{11}$ due to the high-dimensionality of the features and their unbounded range. Moreover, it even increases for most of the adaptation process, contrary to our expectations and what is desired for the method to work. This apparent mystery is elucidated by measuring the same distances at the bottleneck $b_a(s)$. There, as seen in Figure \ref{fig:dist_mse} (right), the euclidean distance behaves as expected during adaptation and the MSE is a reasonable range. Therefore, even if the method itself focuses on the encoder representations, it is the bottleneck that ensures generalization is possible. 

\subsection{Enforcing a Lipschitz constraint}

In this section, we are interested in exploiting the relationship between the smoothness of the actor function and the action-mismatch between the two environments in order to improve the performance in the adaptation environment. Given the high magnitude of the Euclidean distance at the encoder endpoint, it makes sense to enforce a Lipschitz constraint only on the dense layers of the actor that follow the bottleneck.

\begin{wrapfigure}{l}{0.35\textwidth}
    \centering
    \vspace{-8pt}
    \includegraphics[width=\linewidth]{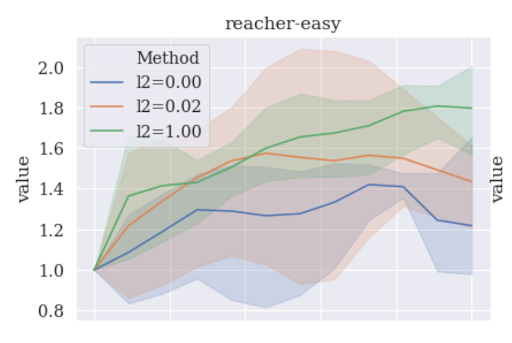}
    \caption{Normalised adaptation performance with Lipschitz constraint. Lower Lipschitz constants correspond to increased generalisation.}
    \label{fig:lipschitz}
    \vspace{-20pt}
\end{wrapfigure}

There is a vast literature on learning Lipschitz continuous functions represented by neural networks \citep{virmaux2018lipschitz, gouk2018regularisation}. Many practical algorithms have been shown to work well in the GAN \citep{goodfellow2014generative} literature and include techniques such as bounding the norm of the gradient \citep{gulrajani2017improved}. In our experiment, we choose the simplest possible method: to reduce the magnitude of the weights with an $L_2$ regularization loss. We train the agent with this additional auxiliary loss weighted by a coefficient $l_2$ and then we adapt it in the target environment as before. 

We show adaption results in Figure \ref{fig:lipschitz} for the reacher-easy environment. As expected, we remark that higher $l_2$ coefficients, corresponding to a smaller $K$, makes the agent increase its adaptation performance from $1.2\times$ to up to $1.8\times$. 

\section{Related Work}

A parallel stream of work has focused on adapting to distractions in the presence of rewards. In this setting, states and observations can be aggregated if they cannot be distinguished with respect to the reward sequences they produce under any action sequences. More generally, \textit{bisimulation} metrics \citep{bisimulation_metric} can be used to quantitatively measure this behavioural similarity. However, they are difficult to compute \citep{10.1137/10080484X, NIPS2008_3423}. Recently, \citet{zhang2020learning} have proposed learning distraction invariant representations by learning an embedding space that respects the bisimulation metric between the observations. Similarly, \citet{pmlr-v97-gelada19a} learn in an unsupervised manner a latent MDP whose norm they theoretically connect to bisimulation metrics. 

In contrast, our work is part of a recent line of research on reward-free adaptation. Closer to the approach we analyse in our paper, \citet{tzeng2017adversarial} use an adversarial procedure to achieve a similar outcome of aligning the features of the two environments by fooling a discriminator that is trained to distinguish between the two. Another class of methods tries to train robust policies by applying various types of domain randomizations \citep{peng2018sim, ramos2019bayessim, tobin2017domain}. While these methods have been successful in making the representations more robust, they cannot possibly anticipate the full set of distractions from a real-world setting.    

\section{Conclusion}

In this work, we analyze the class of self-supervised adaptation methods introduced by \citet{hansen2020self} in a long-term adaptation setting. We propose a geometric picture of the internal process that takes place in the embedding space during adaptation and discover an undesirable aspect of this process: the progressive forgetting of the original representations. We propose a method based on behaviour cloning to fix this problem. Additionally, we quantify the mismatch between actions taken in corresponding states of the two environments and show how it can be reduced further by manipulating the geometry of the actor and critic functions. As a next step, we aim to apply these techniques to real-world robotic applications and distractions specific to these environments \citep{julian2020stop}.

\bibliographystyle{plainnat}
\bibliography{neurips}

\newpage
\appendix

\section{Proofs for Section \ref{sec:theory}}
\label{sec:proofs}

\textbf{Proposition \ref{prop:sac_bound}} \textit{Let $f_a(e) = [\mu(e), \sigma^2(e)]$ be the components of $f_a$ that specify the mean and variance of the multivariate (normal) action distribution of the SAC actor. Additionally, let $\mu$ and $\sigma^2$ be $K$-Lipschitz continuous and $\sigma^2_i \geq \sigma^2_{min}$ for all components $i$. Let $e_1$ and $e_2$ be the embeddings of two matching observations with $d(e_1, e_2) = \eps_e$. Then we have that $\KL(\gN(\mu(e_1), \sigma(e_1)) || \gN(\mu(e_2), \sigma(e_2))) \in O((\eps_e K)^2)$. }

\begin{proof}

We have the KL divergence between two multivariate normal distributions given by
\begin{align}
    \KL = \frac{1}{2}&\Big[ \ln \frac{|\Sigma_2|}{|\Sigma_1|} - D + \Tr(\Sigma_2^{-1}\Sigma_1)  \\ &+ (\mu_2 - \mu_1)^T\Sigma_2^{-1}(\mu_2 - \mu_1) \Big],
\end{align}
where $D$ is the dimension of the random vector. Let $\delta = \max \{\|\mu_1 - \mu_2\|, \|\sigma^2_1 - \sigma^2_2\| \}$. Throughout the proof, we repeatedly use the fact that $\max(\sigma^2_{1,i}, \sigma^2_{2,i}) \leq \min(\sigma^2_{1,i}, \sigma^2_{2,i}) + \delta$ since the Euclidean distance upper-bounds the difference between the individual components of the vectors. 

We begin by bounding each of the terms in this expression by a function of $K$ and $\eps_e$. We start by bounding the logarithm.
\begin{align}
    \ln \frac{|\Sigma_2|}{|\Sigma_1|} &= \ln \prod_i \frac{\sigma^2_{2, i}}{\sigma^2_{1, i}} \leq \ln \prod_i \frac{\min(\sigma^2_{1, i}, \sigma^2_{2, i}) + \delta}{\min(\sigma^2_{1, i}, \sigma^2_{2, i})} \\ 
    &= \sum_i \ln \Bigg(1 + \frac{\delta}{\min(\sigma^2_{1, i}, \sigma^2_{2, i})} \Bigg)
    \leq \sum_i \ln \Bigg(1 + \frac{\delta}{\sigma^2_{min}}\Bigg)
    \\
    &\leq \sum_i \frac{\delta}{\sigma^2_{min}}  \quad \big(\text{since} \ln(1 + x) \leq x, x > -1 \big) \\
    &= \frac{D}{\sigma^2_{min}} \delta 
    \leq \frac{D}{\sigma^2_{min}} K \eps_e \in O(K \eps_e) \quad \big(\text{by Lipschitz continuity} \big)
\end{align}

We can obtain a similar bound for the trace term.
\begin{align}
   \Tr(\Sigma_2^{-1}\Sigma_1) &= \sum_i \frac{\sigma^2_{1, i}}{\sigma^2_{2, i}} \leq \sum_i \frac{\min(\sigma^2_{1, i}, \sigma^2_{2, i}) + \delta}{\min(\sigma^2_{1, i}, \sigma^2_{2, i})} = D + \sum_i \frac{\delta}{\min(\sigma^2_{1, i}, \sigma^2_{2, i})} \\
   &\leq D + D \frac{\delta}{\sigma^2_{min}} \leq D + D \frac{K \eps_e}{\sigma^2_{min}} \in O(K \eps_e) \quad \big(\text{by Lipschitz continuity} \big)
\end{align}

Finally, we bound the last term of the KL divergence:
\begin{align}
&(\mu_2 - \mu_1)^T\Sigma_2^{-1}(\mu_2 - \mu_1) \\
    &\leq \frac{1}{\sigma^2_{min}}(\mu_2 - \mu_1)^T(\mu_2 - \mu_1) \leq \frac{1}{\sigma^2_{min}} \delta^2 \\
    &\leq \frac{1}{\sigma^2_{min}}(K\eps_e)^2 \in O((K\eps_e)^2) \quad \big(\text{by Lipschitz continuity} \big)
\end{align}
Putting it all together, we have $\KL \in O((K\eps_e)^2)$.
\end{proof}

\textbf{Proposition \ref{prop:qtopt_order}} \textit{Let $|f_c(e_1, a_1) - f_c(e_1, a_2)| = \Delta_1$ be the predicted Q-value difference for actions $a_1$ and $a_2$ at observation embedding $e_1$ with $f_c(e_1, a_1) > f_c(e_1, a_2)$. Let $e_2$ be the embedding of another observation. Assume we have a metric over $\gE \times \gA$ with the property that $d([e_1, a_1], [e_2, a_1]) = d([e_1, a_2], [e_2, a_2]) = d(e_1, e_2) = \eps_e$. Additionally, let $f_c$ be $K$-Lipschitz continuous. Then if $\eps_e < \Delta_1/ (2K)$, the order between predicted $Q$ values at $e_2$ is preserved and we have $f_c(e_2, a_1) > f_c(e_2, a_2)$. }

\begin{proof}
Let $|f_c(e_1, a_1) - f_c(e_2, a_1)| = \Delta_2$ and $|f_c(e_1, a_2) - f_c(e_2, a_2)| = \Delta_3$. Then if $\Delta_1 > \Delta_2 + \Delta_3$, the order is preserved, since the summed variation in the two predicted $Q$ values at the state $s_2$ compared to $s_1$ is insufficient to change the order between the two. Using the Lipschitz property of the $Q$ function $f_c$, we can require a stronger inequality to be satisfied:
\begin{align}
    \Delta_2 + \Delta_3 &\leq K\big[ d([e_1, a_1], [e_2, a_1]) + d([e_1, a_2], [e_2, a_2])\big] \quad \big(\text{by Lipschitz continuity} \big) \\ 
    &= 2K\eps_e < \Delta_1
\end{align}
From this, it follows that the order between $Q$ values is preserved if $\eps_e < \frac{\Delta_1}{2K}$
\end{proof}

\textbf{Proposition \ref{prop:qtopt_unimodal}} \textit{Let $d$ be a metric with $d([e_1, a_1], [e_1, a_2]) = d([e_2, a_1], [e_2, a_2]) = d(a_1, a_2) = \eps_a$. Then $\Delta_1 / (2K)$ can be at most $\frac{\eps_a}{2}$}

\begin{proof}
This follows directly from the Lipschitz continuity of $f_c$ and we have 
\begin{equation}
    \eps_e < \frac{\Delta_1}{2K} \leq \frac{K d([e_1, a_1], [e_1, a_2])}{2K} = \frac{1}{2} \eps_a
\end{equation}
\end{proof}

\section{Additional Results}

\subsection{Distance Minimization}
\label{sec:extra_dist}

\begin{figure}[ht]
     \centering
     \begin{subfigure}[b]{0.48\textwidth}
         \centering
         \includegraphics[width=\textwidth]{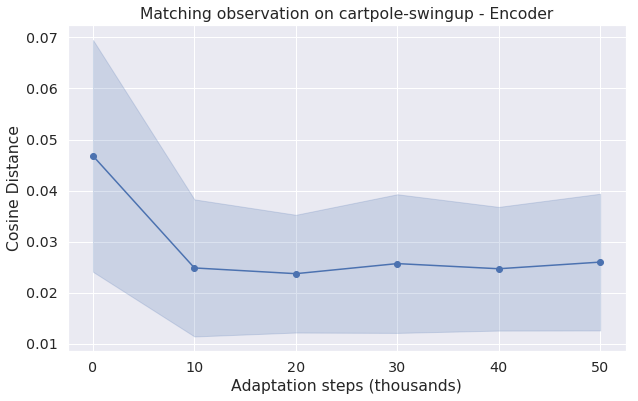}
     \end{subfigure}
     \hfill
     \begin{subfigure}[b]{0.48\textwidth}
         \centering
         \includegraphics[width=\textwidth]{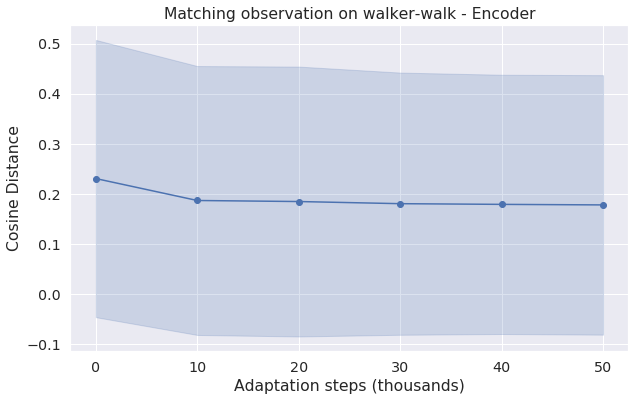}
     \end{subfigure}
     \caption{Cosince distance between encoder emeddings of matching states in the cartpole swingup and walker environments. The matching states of the two environments move towards each other during the adaptation process.}
     \label{fig:extra_match_dist_cos}
\end{figure}

\begin{figure}[h]
     \centering
     \begin{subfigure}[b]{0.32\textwidth}
         \centering
         \includegraphics[width=\textwidth]{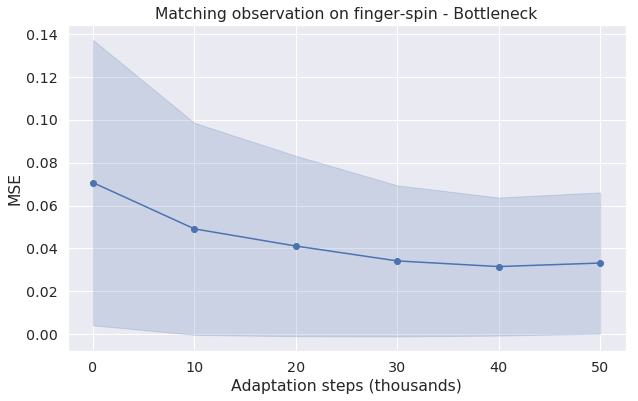}
     \end{subfigure}
     \hfill
     \begin{subfigure}[b]{0.32\textwidth}
         \centering
         \includegraphics[width=\textwidth]{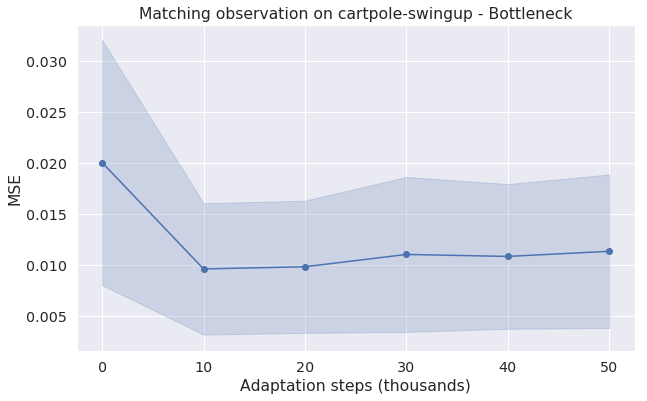}
     \end{subfigure}
     \hfill
     \begin{subfigure}[b]{0.32\textwidth}
         \centering
         \includegraphics[width=\textwidth]{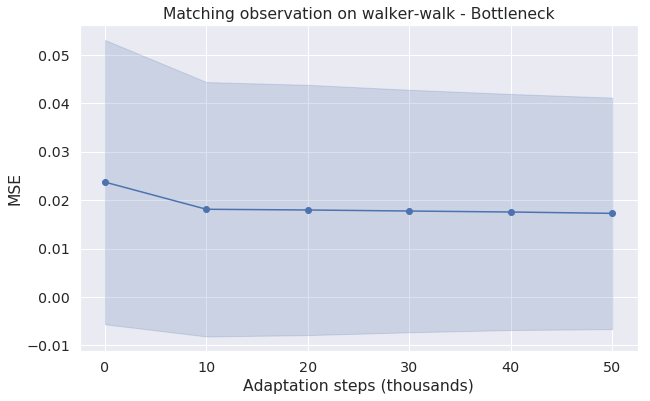}
     \end{subfigure}
     \caption{Distance between bottleneck emeddings of matching states in finger, cartpole and walker environments.}
     \label{fig:extra_dist_mse}
\end{figure}

\subsection{Source environment performance}
\label{sec:extra_source}

\begin{figure}[ht]
    \centering
    \includegraphics[width=\textwidth]{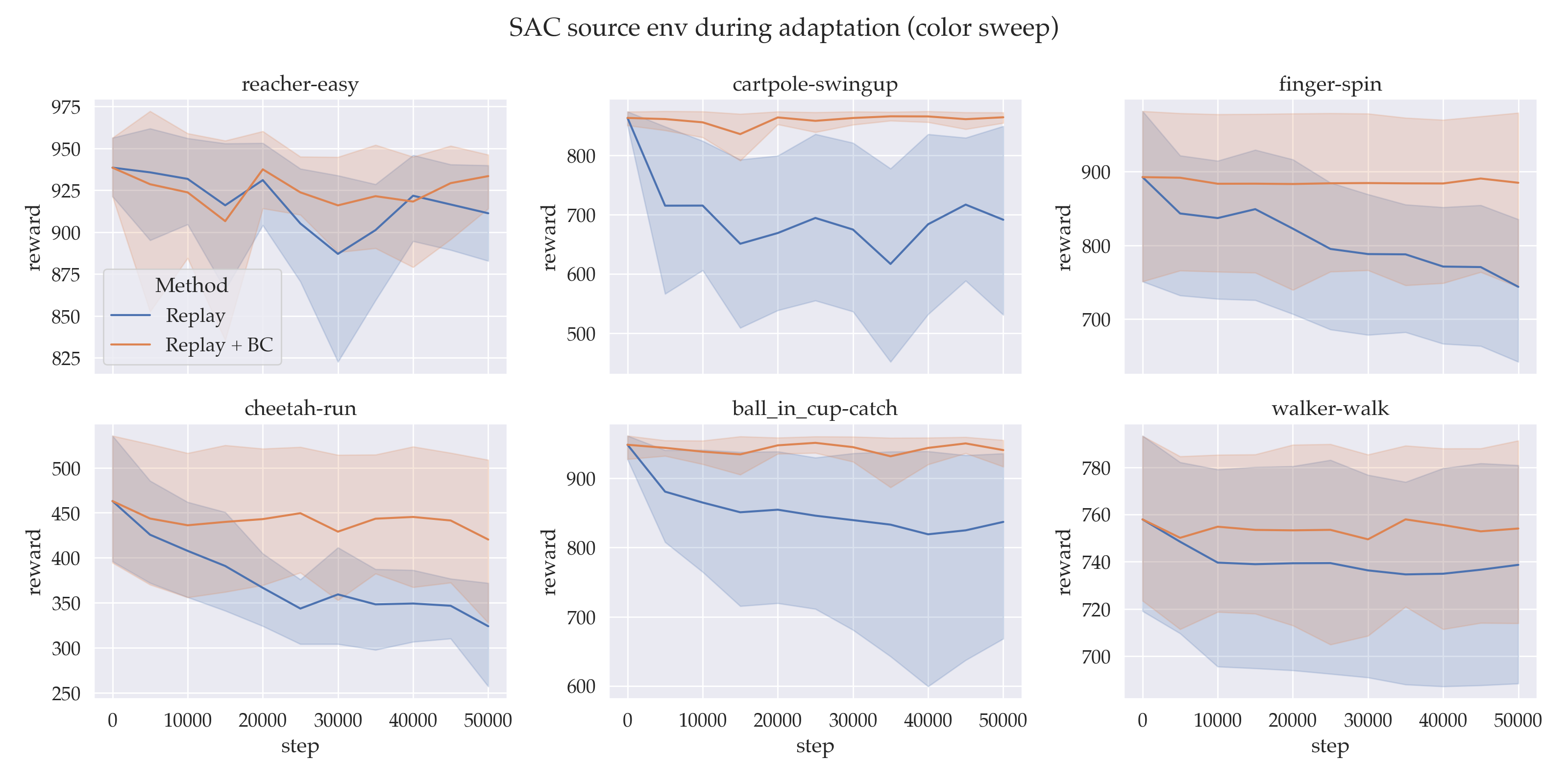}
    \caption{SAC performance in the source environments during colour sweep adaptation. Behaviour cloning generally attenuates the catastrophic forgetting of the policy obtained from the training phase and performance stays constant.}
    \label{fig:extra_sac_source_color}
\end{figure}

\begin{figure}[ht]
    \centering
    \includegraphics[width=\textwidth]{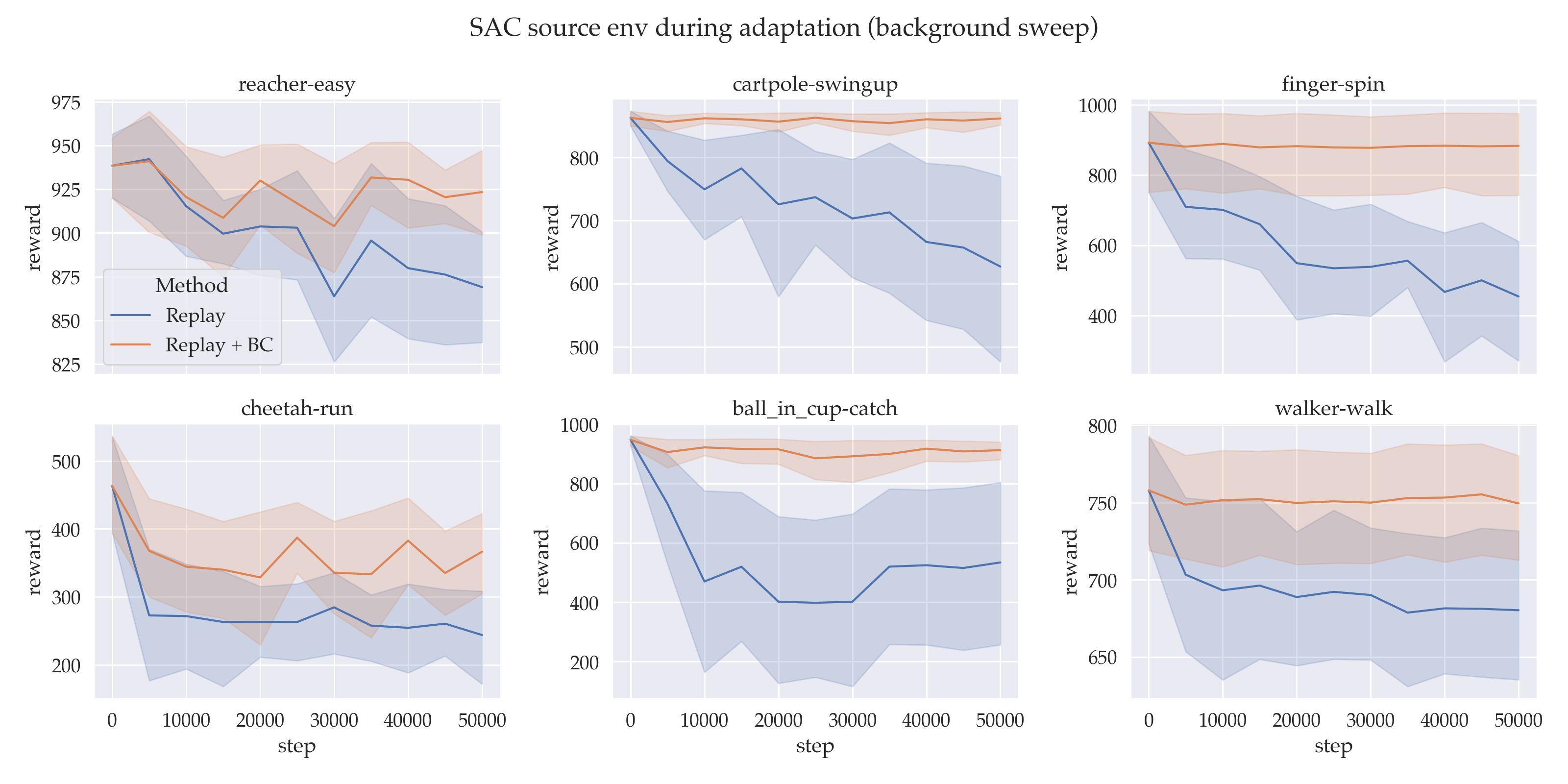}
    \caption{SAC performance in the source environments during background sweep adaptation. Behaviour cloning generally attenuates the catastrophic forgetting of the policy obtained from the training phase and performance stays constant.}
    \label{fig:extra_sac_source_background}
\end{figure}

\begin{figure}[ht]
    \centering
    \includegraphics[width=\textwidth]{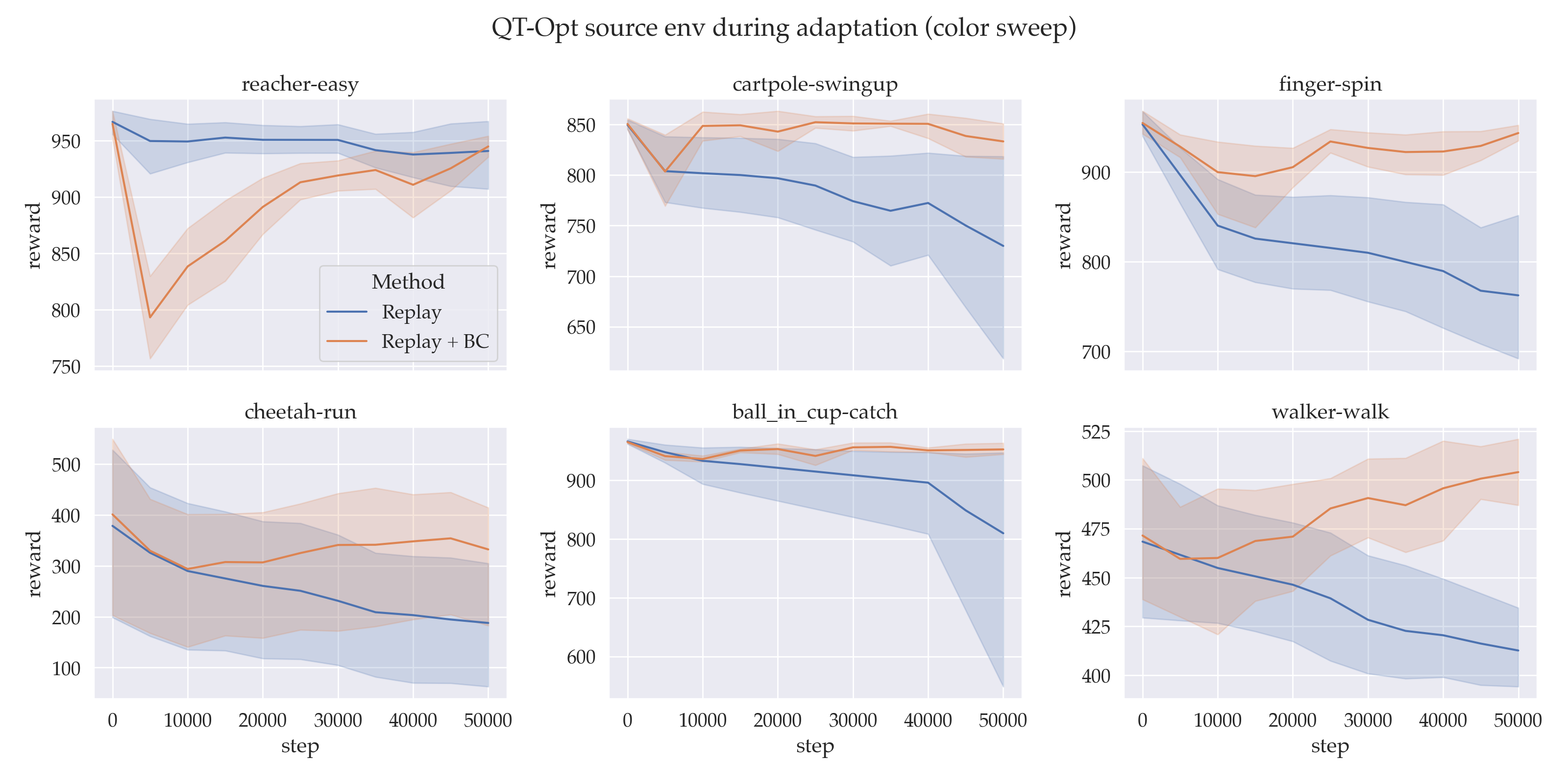}
    \caption{QT-Opt performance in the source environments during colour sweep adaptation. Behaviour cloning generally attenuates the catastrophic forgetting of the policy obtained from the training phase and performance stays constant.}
    \label{fig:extra_qtopt_source_color}
\end{figure}

\begin{figure}[ht]
    \centering
    \includegraphics[width=\textwidth]{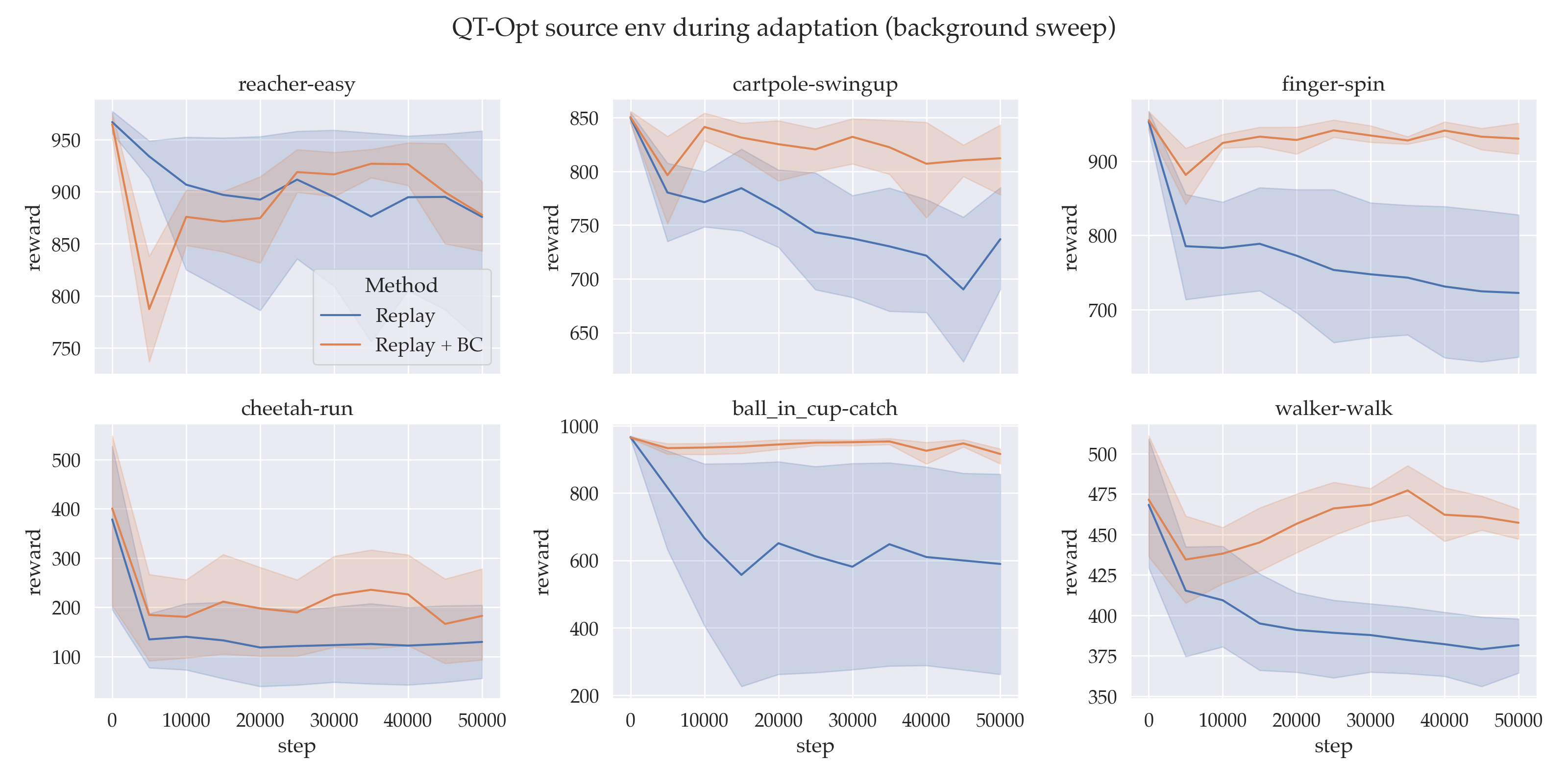}
    \caption{QT-Opt performance in the source environments during background sweep adaptation. Behaviour cloning generally attenuates the catastrophic forgetting of the policy obtained from the training phase and performance stays constant.}
    \label{fig:extra_qtopt_source_bg}
\end{figure}

We include the source environment performance for SAC adapting to color distractions (Figure \ref{fig:extra_sac_source_color}), SAC adapting to background distractions (Figure \ref{fig:extra_sac_source_background}), QT-Opt adapting to color distractions (Figure \ref{fig:extra_qtopt_source_color}) and QT-Opt adapting to background distractions (Figure \ref{fig:extra_qtopt_source_bg}). Across all settings, behaviour cloning attenuates or eliminates completely the effects of catastrophic forgetting.  

\subsection{Adaptation performance}
\label{sec:extra_adapt}

\begin{figure}[ht]
    \centering
    \includegraphics[width=\textwidth]{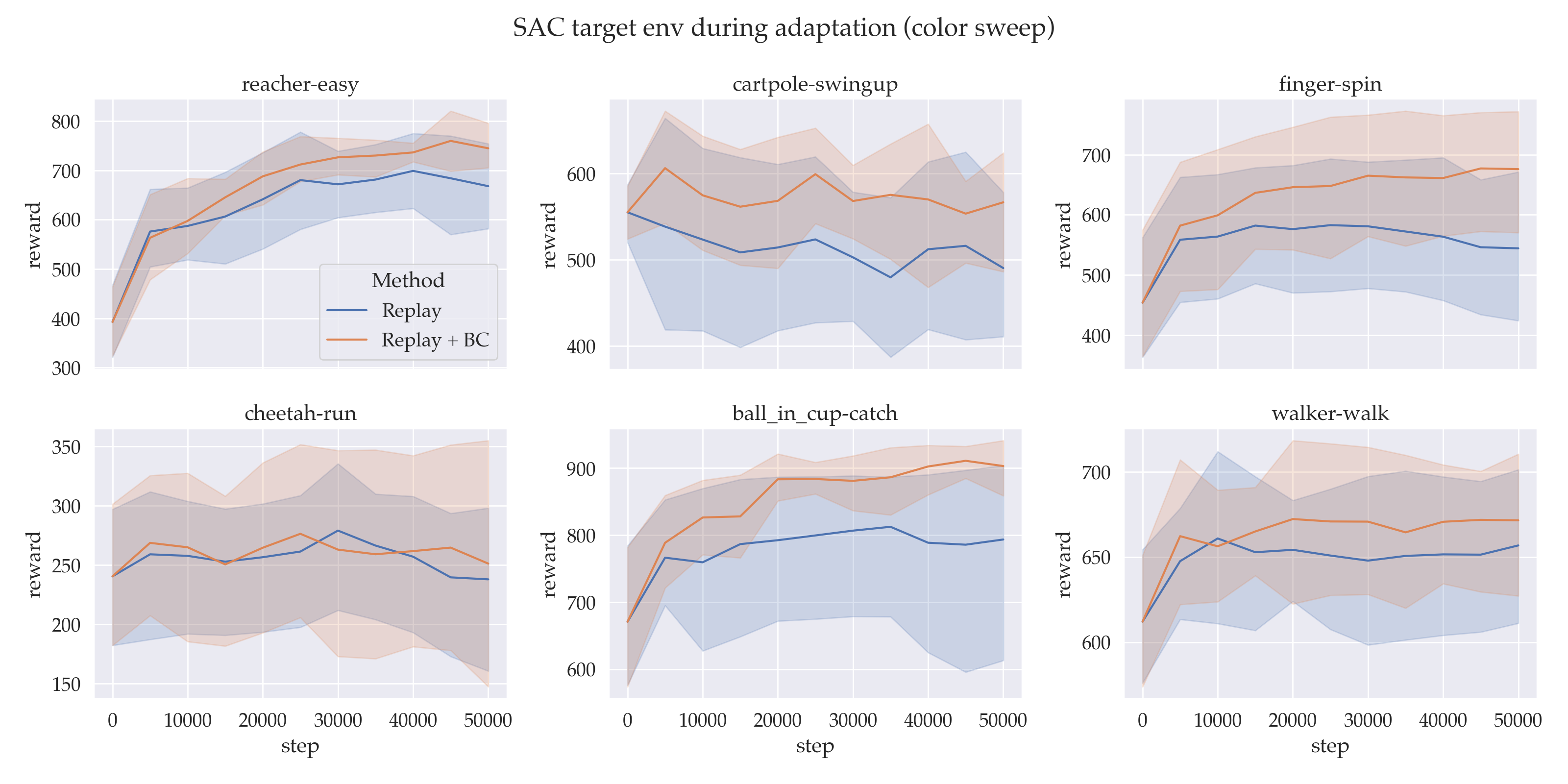}
    \caption{SAC adaptation results for the colour sweep environment. Behaviour cloning either improves or displays the same performance with the exception of the cheetah environment.}
    \label{fig:extra_sac_adapt_color}
\end{figure}

\begin{figure}[ht]
    \centering
    \includegraphics[width=\textwidth]{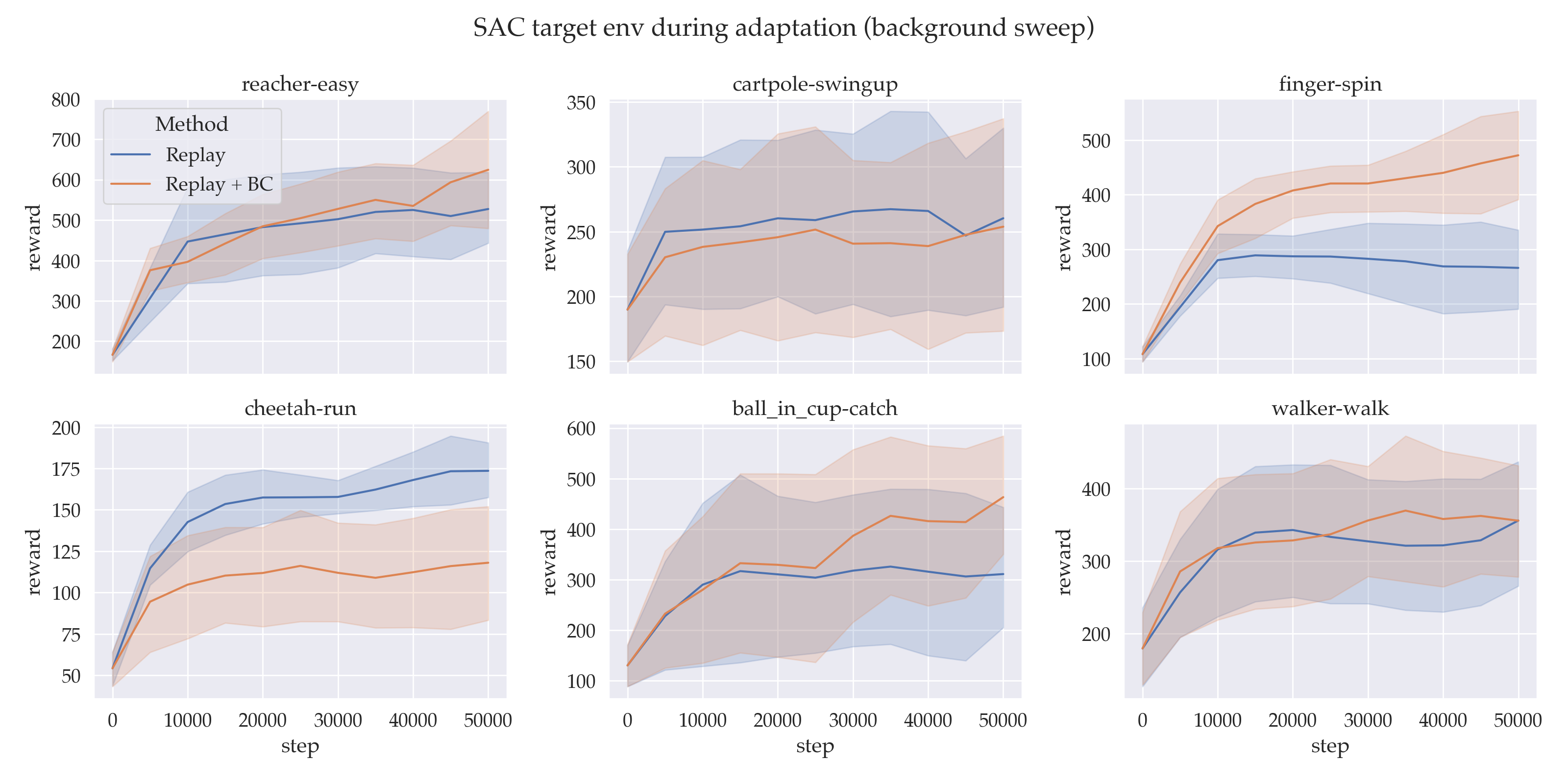}
    \caption{SAC adaptation results for the background sweep environment. Behaviour cloning either improves or displays the same performance with the exception of the cheetah environment.}
    \label{fig:extra_sac_adapt_background}
\end{figure}

\begin{figure}[ht]
    \centering
    \includegraphics[width=\textwidth]{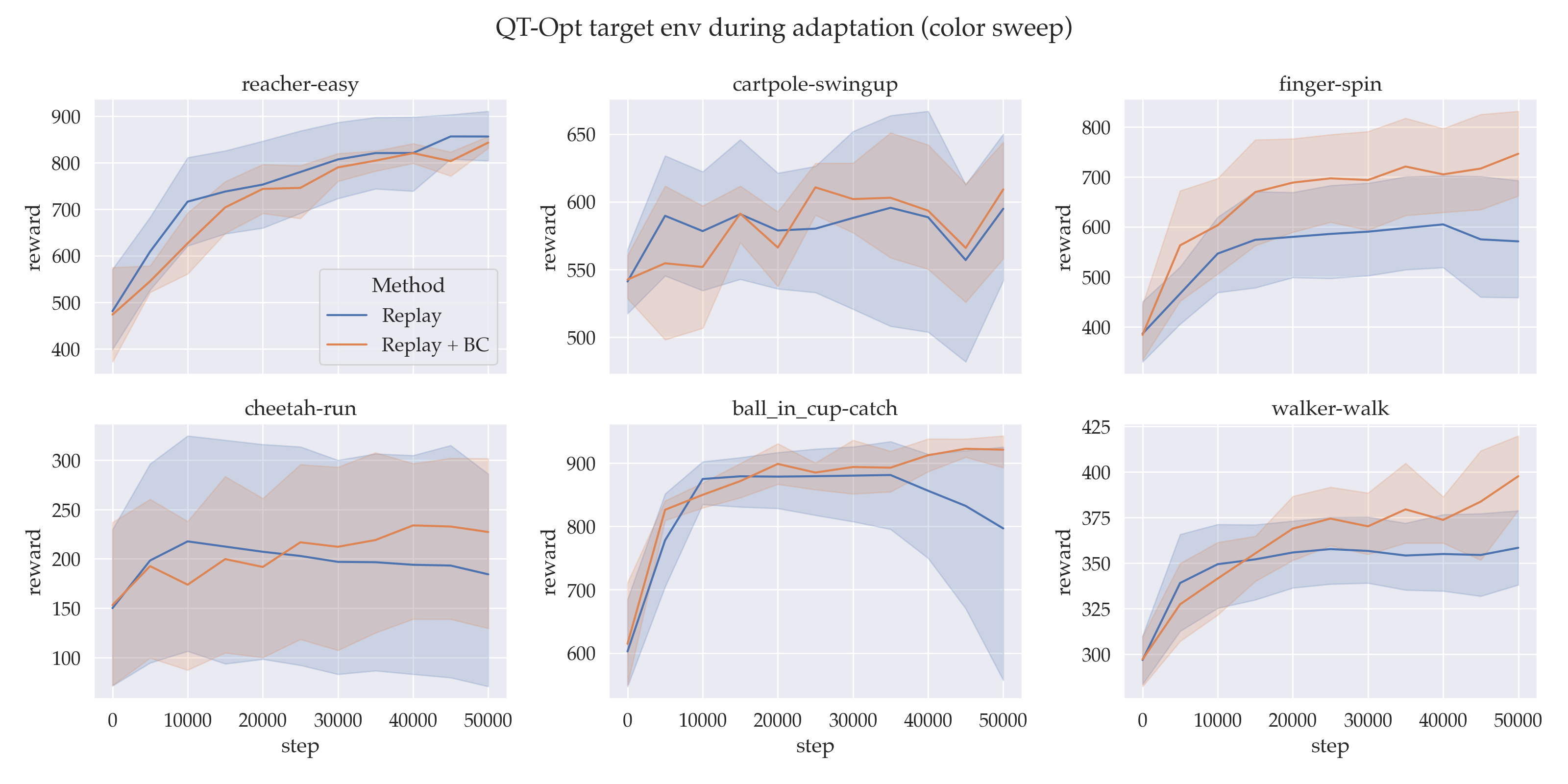}
    \caption{QT-Opt adaptation results for the colour sweep environment. Behaviour cloning either improves or displays the same performance with the exception of the cheetah environment.}
    \label{fig:extra_qtopt_adapt_color}
\end{figure}

\begin{figure}[ht]
    \centering
    \includegraphics[width=\textwidth]{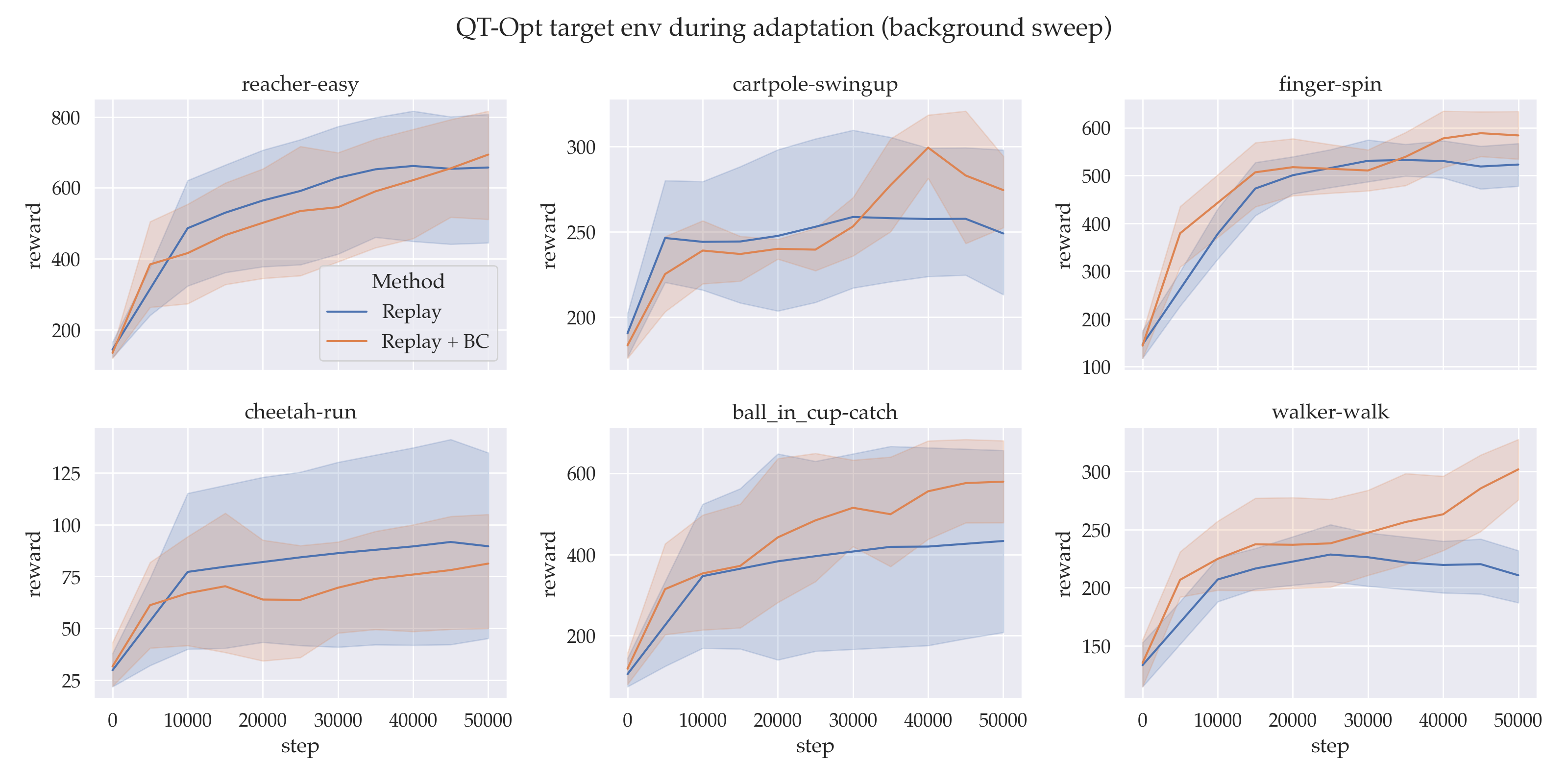}
    \caption{QT-Opt adaptation results for the background sweep environment. Behaviour cloning either improves or displays the same performance with the exception of the cheetah environment.}
    \label{fig:extra_qtopt_adapt_bg}
\end{figure}

We include the target environment performance for SAC adapting to color distractions (Figure \ref{fig:extra_sac_adapt_color}), SAC adapting to background distractions (Figure \ref{fig:extra_sac_adapt_background}, Table \ref{tab:sac_adapt_bg}), QT-Opt adapting to color distractions (Figure \ref{fig:extra_qtopt_adapt_color}, Table \ref{tab:qtopt_adapt_color}) and QT-Opt adapting to background distractions (Figure \ref{fig:extra_qtopt_adapt_bg}, Table \ref{tab:qtopt_adapt_bg}). Behaviour cloning generally improves across almost all combinations of methods and environments.  

\begin{table}[h]
    \centering
    \caption{SAC adaptation in the background distraction environments. Behaviour cloning (BC) either performs similarly or better than the vanilla adaptation.}
    \resizebox{\columnwidth}{!}{
    \begin{tabular}{l | cccccc}
        \toprule
        Method & Reacher-easy & Cartpole-swingup & Finger-spin & Cheetah-run & Ball-in-cup catch & Walker-Walk   \\
        \midrule
        Online & 122.12 $\pm$ 24.43 & 97.60 $\pm$ 3.74 & 21.24 $\pm$ 16.94 & 28.87 $\pm$ 13.74 & 121.56 $\pm$ 28.70 & 42.79 $\pm$ 16.75 \\
        Replay & 527.40 $\pm$ 51.11 & \textbf{260.28 $\pm$ 38.94} & 265.82 $\pm$ 41.94 & \textbf{173.55 $\pm$ 9.96} & 310.83 $\pm$ 67.42 & \textbf{355.68 $\pm$ 46.89} \\
        Replay + BC     & \textbf{624.60 $\pm$ 85.84} & \textbf{253.76 $\pm$ 47.59} & \textbf{471.85 $\pm$ 47.20} & 118.00 $\pm$ 20.81 & \textbf{463.19 $\pm$ 69.65} & \textbf{355.45 $\pm$ 44.38} \\
        \midrule
        RL on Target & 93.00 $\pm$ 10.54 & 192.13 $\pm$ 19.07 & 184.36 $\pm$ 51.53 & 192.13 $\pm$ 19.07 & 78.64 $\pm$ 6.74 & 250.10 $\pm$ 41.53 \\
        Original & 166.26 $\pm$ 8.91 & 189.80 $\pm$ 25.50 & 108.00 $\pm$ 8.18 & 54.20 $\pm$ 6.01 & 129.92 $\pm$ 23.58 & 179.17 $\pm$ 29.81 \\
        \bottomrule
    \end{tabular}
    }
    \label{tab:sac_adapt_bg}
\end{table}

\begin{table}[h]
    \centering
    \caption{QT-Opt adaptation in the color distraction environments. Behaviour cloning (BC) either performs similarly or better than the vanilla adaptation.}
    \resizebox{\columnwidth}{!}{
    \begin{tabular}{l | cccccc}
        \toprule
        Method & Reacher-easy & Cartpole-swingup & Finger-spin & Cheetah-run & Ball-in-cup catch & Walker-Walk   \\
        \midrule
        Online & 468.90 $\pm$ 138.63 & 116.10 $\pm$ 15.88 & 240.03 $\pm$ 39.18 & 66.06 $\pm$ 21.67 & 270.39 $\pm$ 95.18 & 87.04 $\pm$ 29.31  \\
        Replay & \textbf{855.81 $\pm$ 31.08} & \textbf{594.91 $\pm$ 32.96} & 570.85 $\pm$ 72.38 & \textbf{184.30 $\pm$ 64.93} & \textbf{830.78 $\pm$ 56.30} & 358.25 $\pm$ 11.61 \\
        Replay + BC     & \textbf{842.74 $\pm$ 7.45} & \textbf{609.19 $\pm$ 24.79} & \textbf{746.45 $\pm$ 49.33} & \textbf{227.09 $\pm$ 51.75} & \textbf{862.53 $\pm$ 41.25} & \textbf{397.48 $\pm$ 11.41} \\
        \midrule
        RL on Target & 873.50 $\pm$ 17.48 & 556.47 $\pm$ 44.09 & 591.53 $\pm$ 66.04 & 197.33 $\pm$ 63.56 & 825.32 $\pm$ 57.54 & 362.01 $\pm$ 12.16 \\
        Original & 481.16 $\pm$ 52.45 & 541.13 $\pm$ 13.16 & 385.83 $\pm$ 35.32 & 150.11 $\pm$ 42.54 & 490.43 $\pm$ 116.99 & 296.68 $\pm$ 7.56 \\
        \bottomrule
    \end{tabular}
    }
    \label{tab:qtopt_adapt_color}
\end{table}

\begin{table}[!h]
    \centering
    \vspace{5pt}
    \caption{QT-Opt adaptation in the background distraction environments. Behaviour cloning (BC) either performs similarly or better than the vanilla adaptation.}
    \resizebox{\columnwidth}{!}{
    \begin{tabular}{l | cccccc}
        \toprule
        Method & Reacher-easy & Cartpole-swingup & Finger-spin & Cheetah-run & Ball-in-cup catch & Walker-Walk   \\
        \midrule
        Online & 197.57 $\pm$ 55.10 & 89.51 $\pm$ 5.48 & 174.29 $\pm$ 77.61 & 109.47 $\pm$ 23.58 & 109.55 $\pm$ 14.82 & 75.83 $\pm$ 23.74  \\
        Replay & \textbf{657.22 $\pm$ 104.57} & \textbf{249.04 $\pm$ 24.88} & 523.18 $\pm$ 27.06 & \textbf{89.57 $\pm$ 26.60} & \textbf{357.04 $\pm$ 109.15} & 210.67 $\pm$ 12.58 \\
        Replay + BC     & \textbf{693.68 $\pm$ 90.79} &  \textbf{274.48 $\pm$ 12.11} & \textbf{584.23 $\pm$ 29.78} & \textbf{81.21 $\pm$ 16.54} & \textbf{437.40 $\pm$ 84.99} & \textbf{301.87 $\pm$ 15.15} \\
        \midrule
        RL on Target & 695.34 $\pm$ 119.33 & 261.46 $\pm$ 17.55 & 542.61 $\pm$ 33.31 & 84.90 $\pm$ 24.07 & 424.07 $\pm$ 119.80 & 215.25 $\pm$ 16.45 \\
        Original & 143.20 $\pm$ 12.80  & 190.50 $\pm$ 7.18 & 146.25 $\pm$ 16.46 & 29.77 $\pm$ 4.61 & 92.41 $\pm$ 20.14 & 133.23 $\pm$ 11.07  \\
        \bottomrule
    \end{tabular}
    }
    \label{tab:qtopt_adapt_bg}
\end{table}

\end{document}